\documentclass[conference]{IEEEtran}
\IEEEoverridecommandlockouts
\usepackage[T1]{fontenc}
\usepackage{cite}
\usepackage{amsmath,amssymb,amsfonts}
\usepackage{algorithm}
\usepackage{algpseudocode}
\usepackage{graphicx}
\usepackage{subcaption}
\usepackage{booktabs}
\usepackage{textcomp}
\usepackage{xcolor}
\usepackage{cite}
\usepackage{amsmath,amssymb,amsfonts}
\usepackage{graphicx}
\usepackage{textcomp}
\usepackage{xcolor}
\usepackage{graphicx}
\usepackage{etoolbox}
\usepackage{caption}
\usepackage{soul}
\usepackage{float}

\newcommand{\linebreakand}{%
  \end{@IEEEauthorhalign}
  \hfill\mbox{}\par
  \mbox{}\hfill\begin{@IEEEauthorhalign}
}
\newcommand{\insertfig}
{ \setcounter{figure}{0}\includegraphics[width=0.85\textwidth]{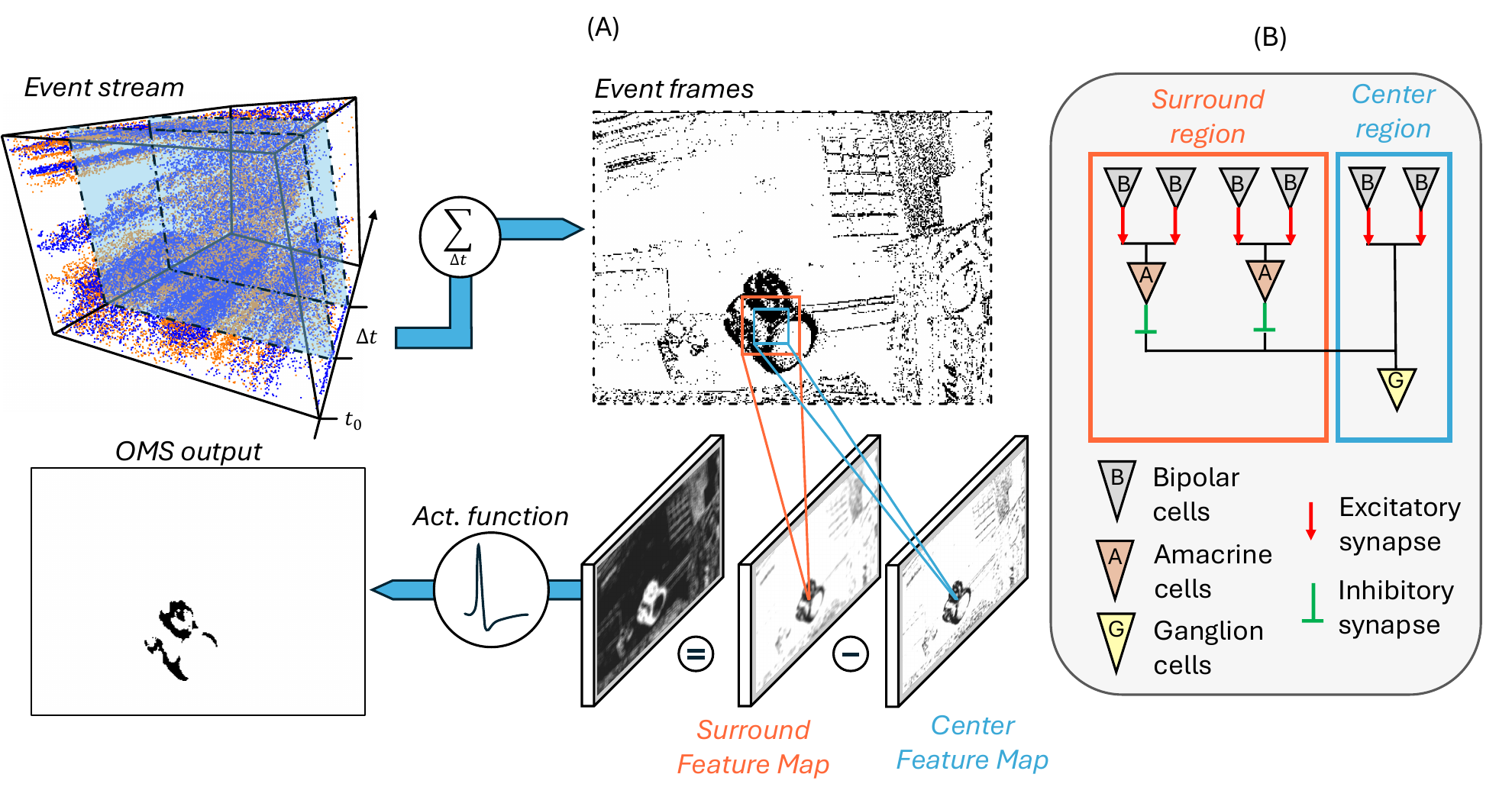}\captionof{figure}{\small (A) An event stream sliced over a time window $\Delta t$ for event frame generation by time accumulation. Event frames serve as the bipolar cell output and input to the Object Motion Sensitivity (OMS) algorithm. Two concentric convolutional filters are depicted in orange and blue representing the surround and center region, respectively. Feature maps obtained from the two convolutions are subtracted and compared to a threshold in order to compute OMS. (B) Retina-inspired OMS biological circuit of surround and center regions of the visual receptive fields.}\label{fig:intro}}

\makeatletter
\apptocmd{\@maketitle}{\centering\insertfig}{}{}

\makeatother

\begin{document}
\title{Retina-Inspired Object Motion Segmentation for Event-Cameras
}
\author{Victoria Clerico\textsuperscript{1,2},
Shay Snyder\textsuperscript{1}, Arya Lohia\textsuperscript{1},
Md Abdullah-Al Kaiser\textsuperscript{3},\\
Gregory Schwartz\textsuperscript{4}, Akhilesh Jaiswal\textsuperscript{3}, Maryam Parsa\textsuperscript{1}
\\

\IEEEauthorblockA{
        \textsuperscript{1} George Mason University, USA \\
        \textsuperscript{2} IBM Research Zürich, Switzerland \\
        \textsuperscript{3}
        University of Wisconsin-Madison, USA \\
        \textsuperscript{4}
        Northwestern University, USA \\
    }
    }
\maketitle

\begin{abstract}
Event-cameras have emerged as a revolutionary technology with a high temporal resolution that far surpasses standard active pixel cameras. This technology draws biological inspiration from photoreceptors and the initial retinal synapse. This research showcases the potential of additional retinal functionalities to extract visual features. We provide a domain-agnostic and efficient algorithm for ego-motion compensation based on Object Motion Sensitivity (OMS), one of the multiple features computed within the mammalian retina. We develop a method based on experimental neuroscience that translates OMS' biological circuitry to a low-overhead algorithm to suppress camera motion bypassing the need for deep networks and learning. Our system processes event data from dynamic scenes to perform pixel-wise object motion segmentation using a real and synthetic dataset. This paper introduces a bio-inspired computer vision method that dramatically reduces the number of parameters by $\text{10}^\text{3}$ to $\text{10}^\text{6}$ orders of magnitude compared to previous approaches. Our work paves the way for robust, high-speed, and low-bandwidth decision-making for in-sensor computations. 
\end{abstract}


\section{Introduction}
\label{sec:intro}
\noindent
Digital cameras have become essential tools for capturing visual information in our surroundings. Their applications range from smartphones~\cite{Hayes:12} and autonomous driving~\cite{ma20223d}, to robotics~\cite{snyder2021thor}, and manufacturing~\cite{zhou2022computer}. However, traditional cameras lack dynamic range and temporal resolution. Event cameras represent the next generation of imaging technology by drawing inspiration from biological retinas. They transmit per-pixel intensity changes, unlike the color-focused information captured by RGB cameras. Their design reduces latency and power consumption, provides a high dynamic range, and reduces motion blur, making them ideal for high-speed applications \cite{gallego2020event}.
Event-cameras, such as Dynamic Vision Sensors (DVS), respond asynchronously to brightness changes in the scene and generate a burst of binary events containing space-time coordinates~\cite{gallego2020event}. This sparse representation of spatiotemporal activity allows addressing complex computer vision tasks such as optical flow estimation \cite{bartolozzi2014flow, stroffregen2018flow} and motion segmentation \cite{stroffregen2019, gallego2018unifying, EED}. 

In applications where event cameras are mounted on moving platforms, distinguishing events caused by moving objects (object motion) from those caused by camera shifts (ego-motion) is an ongoing challenge \cite{khan2017ego}. Numerous methods address this task by employing optimization algorithms or neural networks to compensate for ego-motion \cite{zhu2019unsupervised, GConv, multi-rnn}. However, these methods carry a large number of parameters and rely on extensive training, making them impractical for real-time applications.

As an alternative to computationally intensive methods, a bio-inspired solution arises from the neural circuitry of the animal retina, known as Object Motion Sensitivity (OMS). OMS is implemented through feature-spike activity within retinal ganglion cells (RGCs)-the output cells of the retina-to discriminate external motion from the eye's motion through center-surround suppression \cite{lveczky2003SegregationOO, baccus2008retinal}. The center-surround difference in the visual receptive field (RF) has been previously applied to object detection in infrared videos \cite{saliency}. However, this method lacks comprehensive retinal bio-inspiration, bypassing bipolar behavior, and assumes static cameras without addressing ego-motion.
High Dynamic Range sensors and DVS are inspired by computations in the outer retina, i.e. phototransduction and the first retinal synapse. This work delves deeper into the inner retina by focusing on computations performed at RGCs to derive feature-rich information.

Prior work introduced an electronic circuit capable of 3D semiconductor integration for in-sensor OMS computation \cite{yin2022iris}. Most recently, a reconfigurable design was presented to enable re-engineered OMS computations adaptable to various scenarios~\cite{sinaga2024hardwarealgorithmreengineeringretinalcircuit}. This paper leverages OMS visual computation as a bio-inspired algorithm for DVS cameras. The proposed algorithm has similar or improved performance compared to seven state-of-the-art methods for object motion extraction with up to $10^6$x parameter reductions. Unlike methods relying on large post-processing models, OMS low-parameter density allows for direct in-sensor integration and post-processing, adapting to varying resource constraints. This work paves the way for developing more brain-inspired, scalable computer vision systems incorporating multiple visual features derived from experimental neuroscience. The major contributions of this paper are the following:
\begin{enumerate}
 \item We introduce a biologically plausible computer vision algorithm, modeled after the mammalian retina, to compute Object Motion Sensitivity - one among the other 40 fundamental features in the visual system.
\item We demonstrate its direct applicability to DVS, bridging the gap between theoretical neuroscience and real-world computer vision applications.
\item We show that our bio-inspired algorithm surpasses or matches the performance of seven state-of-the-art deep learning approaches while avoiding the complexities of domain-specific training.
\item We achieve a reduction in model complexity, with our method requiring between 3 to 6 orders of magnitude fewer parameters than leading deep learning approaches.
\end{enumerate}

\section{Related Work}
\noindent
To the best of our knowledge, this work serves as a foundational work of end-to-end retinal computations applied to existing computer vision tasks, in particular, motion segmentation (MS). MS is effectively addressed by event-based feature methods with event-cameras on fixed positions, where spikes correspond solely to moving objects~\cite{motion2006, motion2007, motion2012, motion2013}. However, tracking moving objects becomes difficult in scenarios with camera motion (i.e. ego-motion). 

\textbf{Ego-Motion Compensation for Motion Segmentation} is a persistent issue raised by using cameras mounted on moving platforms for target detection. DVS detects pixel-level brightness changes due to platform shifts, generating events for targets and static objects. 
Thus, the spikes generated by the background movement obstruct target objects, complicating their tracking.

Earlier works addressed this problem by introducing algorithmic and parametric models. Stroffregen et al.~\cite{stroffregen2019} proposed a per-event MS method using a clustering algorithm. Mitrokhin et al.~\cite{EED} introduced a parametric model for object tracking compatible with event-based cameras to reduce ego-motion noise. 
Vasco et al. \cite{vasco2017icub} proposed a method to differentiate independent motion from ego-motion and estimate the velocities of salient objects. Traditional methods, while effective, can be computationally expensive. This novel approach takes a unique path by taking inspiration from the biological processes of the mammalian retina to achieve efficient performance. Recent works on ego-motion compensation focus on developing learning algorithms, especially based on deep or shallow neural networks. By analyzing visual inputs such as RGB frames or DVS
streams, these models extract fundamental features for accurate decision-making in lower-dimensional representations. Methods proposed by Mitrokhin et al. \cite{mitrokhin2020evimo} and Nitin et al. \cite{MOD} suggested handling the events as a 3-channel event frame, where the first and second channels are the positive and negative event counts, and the third is the average time between events. This event frame is then propagated through Convolutional Neural Networks (CNNs).

Zhu et al. \cite{zhou2017unsupervised} proposed using bilinear sampling to discretize the time domain. Resultant frames are fed to a CNN autoencoder with a pose estimation block for ego-motion. Zhang et al. \cite{multi-rnn} showed Multi-Scale Recurrent Neural Networks for temporal dependencies capturing on the DVS stream. Furthermore, architectures such as GConv \cite{GConv} explored the events in spatio-temporal 3D space. Additionally, Spiking Neural Networks exemplified by SpikeMS \cite{SpikeMSs}, offered a different solution to leverage the sparsity and time-dependency of the events. These intelligent systems have enhanced capabilities, yet they require extensive learning to overcome noisy and low-entropy visual representations of the
input events. The training process leads to poor domain adaptation and generalization. This paper seeks to establish a novel approach for MS and ego-motion challenge using end-to-end retinal computations, directly applicable to DVS cameras, bypassing the need for domain specific training.

\section{Methodology}
\noindent
The visual pathways of many biological organisms have evolved the OMS circuit to suppress stimuli created by the eye's movements. As shown in Figure \ref{fig:intro}B, this circuit is comprised of bipolar, amacrine, and retinal ganglion cell (RGC), which work together to distinguish between stimuli created by global and local motion. 

Neuroscience foundations define the role these cells play in the OMS computation \cite{schwartz2021object}. Bipolar cells rapidly adapt to luminance changes, transmitting information about both light increases and decreases, which is then refined and conveyed to the brain by RGCs as spike trains through the optic nerve. Clusters of interconnected bipolar cells and retinal ganglion cells (RGCs) respond to light stimuli in the visual receptive fields (RFs)~\cite{Baden2016RGC} and play a crucial role in suppressing the eye's motion. Bipolar and ganglion cells RFs are divided into the center and surround regions. The center region of the RF is formed by direct connections between bipolar cells and ganglion cells, while the surround region results from indirect interactions between both types of cells through amacrine cells \cite{receptive2018Fields}. Thereby, amacrine cells integrate contrast signals from the surround RF and subtract them from the center RF, extracting a \textit{differential motion} between a global and a local area \cite{lveczky2003SegregationOO}.

This computation acts like spatial filtering that detects changes in movement between global and local areas. Therefore, we propose an algorithm, see Algorithm~\ref{oms_neuro}, that approximates this mechanism using convolutional kernels, as they are directly inspired by visual receptive fields.


\begin{algorithm}[h]    
\caption{OMS - Neuroscience}\label{oms_neuro}
\begin{algorithmic}[1]
\Require {radius $r_1$ $<$ radius $r_2$}
\Procedure{OMS}{$dvsFrames$, $r_1$, $r_2$, $s_s$, $\alpha$}
\State $s_c\gets s_s + r_2 - r_1$
\State $center\gets gaussianKernel(r_1)$
\State $surr \gets gaussianKernel(r_2)$

\For{$i$ in $len(dvsFrames)$}{}
    \State $frames \gets dvsFrames[i]$
    \State $fil\_c \gets applyKernel(center, frames, s_c)$
    \State $fil\_s \gets applyKernel(surr, frames, s_s)$
    \State $events \gets |fil\_c - fil\_s$|
    \State $OMS[i] \gets events > \alpha$
\EndFor
\State \textbf{return} $OMS$
\EndProcedure
\end{algorithmic}

\end{algorithm}

\noindent
An in-depth description of each parameter in our OMS algorithm is described below:
\begin{itemize}
    \item \textbf{Input frames ($dvsFrames$)} store photoreceptor activations within bipolar cells. Our practical approach projects the events into a frame by compressing the time and polarity domain. This input provides the binary values mimicking the bipolar cells output. 
    \item \textbf{Center ($center$) and surround kernels ($surr$)} represent two circular Gaussian kernels of a chosen radius ($r_1$ and $r_2$), with discrete values sampled from normalized distributions. 
    \item \textbf{Radius ($r_1$ and $r_2$)} represent the radii that define the discrete circles in the filters' matrices.
    \item \textbf{Strides of the center ($s_c$) and surround ($s_s$) kernels}.
    \item \textbf{Threshold} ($\alpha$) decides whether motion is detected for a pixel.
\end{itemize}
DVS frames (\textit{dvsFrames}) represent the photoreceptor activations within bipolar cells, and OMS performs the computations of the amacrine and RGC layers. The algorithm consists of two averaging filters, containing a discrete feathered circle of a chosen radius whose values sum to one. The values in the center of the matrix are larger and therefore carry more significance.


The smaller of these kernels is the center kernel ($center$), which represents the RGCs and bipolar cell cluster to which it is connected, see Figure \ref{fig:intro}B. We set its radius $r_1 = 2$ to reduce the likelihood that a single cell will cover an entire object. The surround kernel ($surr$) serves as the amacrine cell and its corresponding bipolar cell cluster, which is designed to inhibit the RGCs' response if global motion is perceived. For this kernel, $r_2=4$ was chosen to cover a sufficiently wide region of each frame. The surround kernel convolves over the frame with a stride of $s_s = 1$ and the center kernel stride is dynamically calculated based on the surround kernel stride with the difference in kernel radius.

Finally, to simulate inhibition, the mean contrast values from the surround kernel are subtracted from those of the center kernel, see inhibitory and excitatory synapses in Figure \ref{fig:intro}B. In case the resultant value is larger than a threshold ($\alpha$) for a given pixel, then a binary spike is stored in the OMS frame. The threshold value was found empirically and set to $\alpha = 0.96$ for EV-IMO~\cite{mitrokhin2020evimo}.
The selected value offers a balance between preserving object structure and suppressing background interference. In contrast, we selected a threshold of $\alpha = 0.5$ for MOD \cite{MOD} because of reduced spike density.

\section{Evaluation}
\noindent
This study addresses object motion segmentation in the presence of ego-motion using our novel OMS algorithm. Through comprehensive experimentation on two datasets, we seek to showcase the superior performance and simplicity of our algorithm compared to state-of-the-art methods.
\subsection{Datasets and Metrics}
\noindent
To simulate the visual phenomenon where OMS is particularly effective, we selected a set of public DVS datasets for motion segmentation. 
\begin{table*}[t]
\centering
\caption{Quantitative results using the mean IoU(\%) metric on EV-IMO \cite{mitrokhin2020evimo} validation set sequences. Results are shown for times between event slices of 100 ms and 20 ms.}
\label{tab:qualitative_EVIMO}
\renewcommand{\arraystretch}{1.3}
\setlength{\tabcolsep}{3pt}
\fontsize{10pt}{10pt}\selectfont
\begin{tabular}{lcccccccccccc}
\multicolumn{1}{c}{Method} & \multicolumn{2}{c}{boxes} & \multicolumn{2}{c}{floor} & \multicolumn{2}{c}{wall} & \multicolumn{2}{c}{table} & \multicolumn{2}{c}{fast}& tabletop  \\ 
\multicolumn{1}{c}{} & 100 & \multicolumn{1}{c}{20} & 100 & \multicolumn{1}{c}{20} & 100 & \multicolumn{1}{c}{20} & 100 & \multicolumn{1}{c}{20} & \multicolumn{1}{c}{100} & \multicolumn{1}{c}{20} \\ \hline
\textit{EV-IMO} \cite{mitrokhin2020evimo}& \multicolumn{2}{c}{70$\pm$5}  & \multicolumn{2}{c}{59$\pm$9}  & \multicolumn{2}{c}{78$\pm$5} &\multicolumn{2}{c}{79$\pm$6}& \multicolumn{2}{c}{67$\pm$3}  \\ \hline
\textit{PointNet++$^{*}$} \cite{pointnet++} & 71$\pm$22& 80$\pm$15 &68$\pm$18 &76$\pm$10&75$\pm$19 &74$\pm$20& 62$\pm$28&68$\pm$23&24$\pm$10&20$\pm$6 \\ \hline
\textit{GConv$^{*}$} \cite{GConv}& 81$\pm$8 & 60$\pm$18 & 79$\pm$7 & 55$\pm$19& 83$\pm$4& 80$\pm$7 & 57$\pm$14 & 51$\pm$16 & 74$\pm$17 & 39$\pm$19 \\ \hline
\textit{SpikeMS$^{*}$} \cite{SpikeMSs}& 61$\pm$7&65$\pm$8&60$\pm$5&53$\pm$16&65$\pm$7&63$\pm$6&52$\pm$13&50$\pm$8&45$\pm$11&38$\pm$10\\ \hline
\textit{MSRNN}\textsuperscript{†} \cite{multi-rnn}& \multicolumn{2}{c}{80}  & \multicolumn{2}{c}{85} & \multicolumn{2}{c}{82} & \multicolumn{2}{c}{85} & \multicolumn{2}{c}{75}  & \multicolumn{2}{c}{87} \\ \hline
Ours (OMS) & \multicolumn{2}{c}{\textbf{72$\pm$16}}& \multicolumn{2}{c}{\textbf{94}} & \multicolumn{2}{c}{\textbf{82$\pm$6}} & \multicolumn{2}{c}{\textbf{88$\pm$10}} & \multicolumn{2}{c}{\textbf{69$\pm$3}} & \multicolumn{2}{c}{\textbf{72$\pm$14}} 
\end{tabular}
\\
\vspace{2pt}
\fontsize{9.5pt}{10.25pt}\selectfont
\textsuperscript{*} Results taken from \cite{SpikeMSs}. \textsuperscript{†} Results taken from \cite{multi-rnn}.
\end{table*}

\textbf{I. EV-IMO}~\cite{mitrokhin2020evimo} is an event-based motion segmentation dataset captured with a DAVIS 346C DVS camera at 200 Hz, featuring $346\times260$ resolution and a 70$^\circ$ field of view. It includes 32 minutes of recordings with up to three moving objects in controlled environments. Each event encodes polarity, coordinates, and timestamp, while ground truth motion masks, generated at 40 Hz using the \textit{VICON motion capture system}, are provided as $346\times260$ binary matrices with associated timestamps. Since DVS events and the motion masks are captured asynchronously and at different rates, we pre-processed and selected a set of events depending on the capturing time of the mask to encourage time correspondence. To ensure a fair comparison, we adopted the preprocessing methodology from \cite{SpikeMSs}, one of the benchmark approaches. For our experiments, DVS events are transformed into frames. The filtered events corresponding to a single motion mask are accumulated over time and the two polarity dimensions are compressed into one. This procedure results in frames of size $346\times260$. 

\vspace{3pt}
\noindent
\textbf{II. MOD}~\cite{MOD}, \textit{Moving Object Dataset}, is a synthetic dataset generated by simulating a 3D room with seven unique objects. The simulated scenes have different background wall textures and object and camera trajectories. The images are rendered at 1000 frames per second, with a resolution of $346\times260$, and a 90$^\circ$ field of view. Then, the events are generated using the approach proposed in~\cite{simulator_MOD}. An identical preprocessing step was applied for the MOD events to provide a correspondence between motion masks and events.
\subsection{Comparisons with state-of-the-art Methods}


\noindent
The quantitative evaluation is performed using the mean Intersection over Union ($mIoU$) and Detection Rate ($DR$) over the validation sequences, as defined in \cite{MOD}. The $mIoU$ quantifies the degree of overlap between the predicted segmentation masks and their corresponding ground truth annotations. 
The $DR$ metric considers a successful prediction if the intersection between the ground truth mask and the prediction mask is at least 50\% of the ground truth area and greater than the intersection with the outside area of the ground truth mask.
 
The prediction masks of the proposed method are the DVS frame masked with the result of the OMS computation. OMS takes as input the projected DVS events with a compressed polarity dimension. In contrast, the ground truth mask corresponds to the DVS event frames masked with the original motion mask. This removes spikes associated with stationary objects while retaining spikes generated by moving objects.


Previous approaches mainly focus on deep or shallow neural networks to leverage DVS data. EV-IMO \cite{mitrokhin2020evimo}, and EvDodgeNet \cite{MOD} introduced architectures based on CNNs trained on DVS frames to address the motion segmentation task. Furthermore,  0-MMS demonstrated a splitting-and-merging clustering approach trained on event clouds. Similarly, Gconv \cite{GConv} and PointNet++ \cite{pointnet++} are 3D CNNs trained on event clouds. Moreover, SpikeMS \cite{SpikeMSs} is a spiking neural network autoencoder trained on the DVS event-stream to generate labels at each time step. Finally, MSRNN \cite{multi-rnn} introduced a multi-scale deep recurrent neural network. Some authors provided the mean IoU(\%) using different temporal windows for frame generation of 100 ms and 20 ms. 

Table \ref{tab:qualitative_EVIMO} shows the results of different architectures trained and evaluated on EV-IMO \cite{mitrokhin2020evimo}. For all validation sequences, the OMS algorithm results in equivalent or better performance, see Table \ref{tab:qualitative_EVIMO}. However, the OMS operates without any form of training, demonstrating a greater degree of determinism and domain independence.  The results of the EV-IMO \cite{mitrokhin2020evimo} and MOD \cite{MOD} validation sets are summarized in Table \ref{tab:mIoU_final} and \ref{tab:dr}, compared to state-of-the-art methods. 

\begin{figure*}[t]
    \centering
    \includegraphics[width=0.825\textwidth]{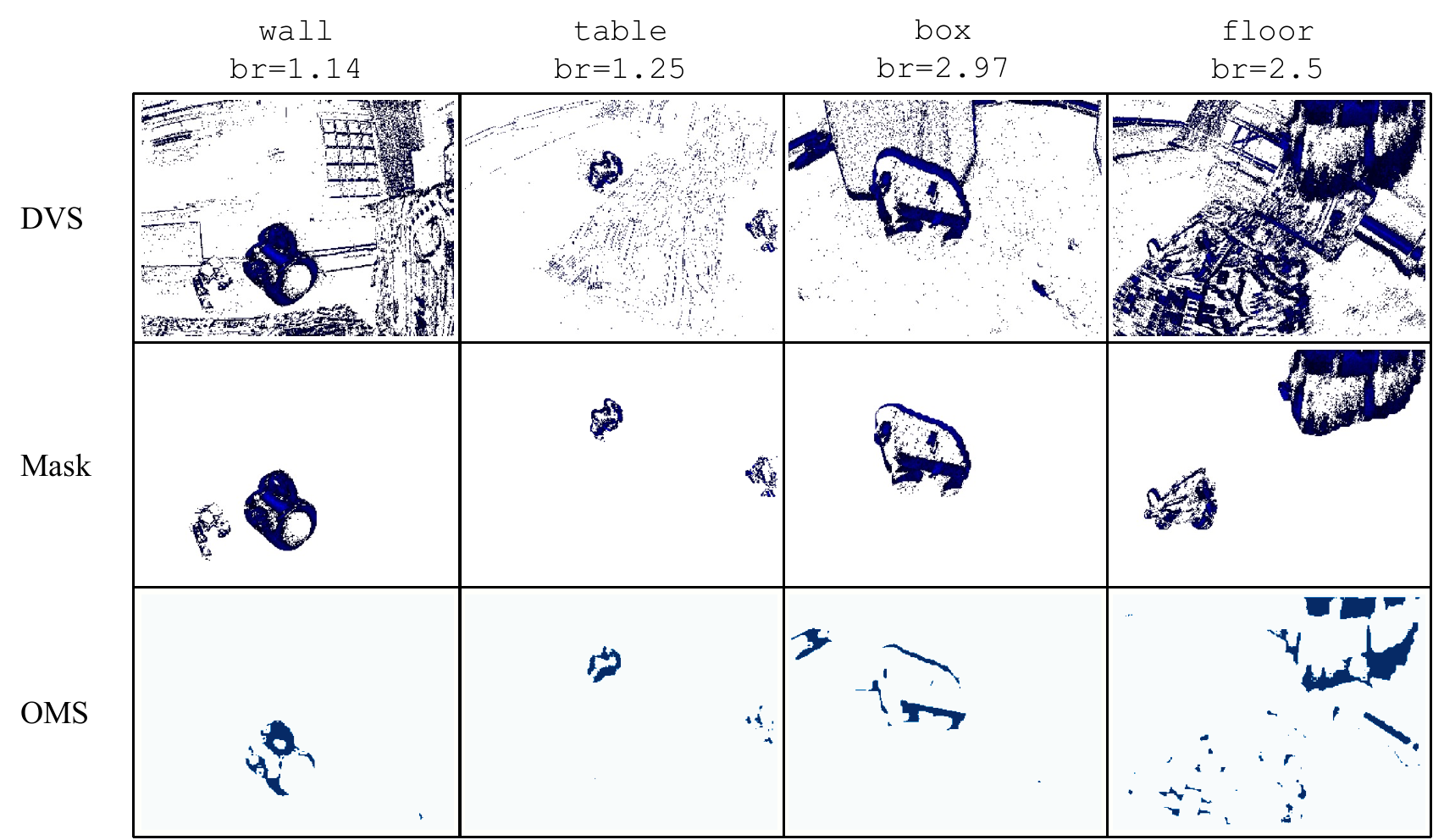}
    \caption{Comparative analysis between DVS event frames, ground truth mask and OMS for table, wall, floor, box, and fast sequences from EV-IMO \cite{mitrokhin2020evimo}.}
    \label{fig:sequences_evimo}
\end{figure*}

\begingroup
\renewcommand{\arraystretch}{1.1}
\begin{table}[h]
\centering
\caption{mIoU (\%) for validation sequences using EV-IMO \cite{mitrokhin2020evimo} and MOD \cite{MOD} datasets.}
\label{tab:mIoU_final}
\fontsize{10pt}{10pt}\selectfont
\begin{tabular}{lcc}

\toprule
Method & EV-IMO \cite{mitrokhin2020evimo} & MOD \cite{MOD}\\ 
\midrule
EV-IMO$^{*}$\cite{mitrokhin2020evimo} & 77 & - \\
EVDodgeNet$^{*}$\cite{EED} & 65.76 & 75 \\
0-MMS$^{*}$ \cite{MOD++} & 80.37 & - \\
SpikeMS \cite{SpikeMSs} & - & 68 \\
Ours (\textit{OMS})  & \textbf{80.07} & \textbf{81.59}\\
\bottomrule
\end{tabular}

\vspace{2pt}
\fontsize{9.5pt}{10.25pt}\selectfont
\textsuperscript{*} Results taken from \cite{MOD++}.
\end{table}
\endgroup

\begingroup
\renewcommand{\arraystretch}{1.1}
\begin{table}[h]
\centering
\caption{Detection Rate (\%) for validation sequences using EV-IMO \cite{mitrokhin2020evimo} and MOD \cite{MOD} datasets.}
\label{tab:dr}
\fontsize{10pt}{10pt}\selectfont
\begin{tabular}{lcc}
\toprule
Method & EV-IMO \cite{mitrokhin2020evimo}  & MOD  \cite{MOD} \\
\midrule
EVDodgeNet$^{*}$\cite{EED}    & 48.79   & 70.12 \\
0-MMS$^{*}$ \cite{MOD++}      & 81.06 & 82.35 \\
SpikeMS \cite{SpikeMSs} & 68.82 & 65.14 \\
\textbf{Ours (\textit{OMS})} & \textbf{92.57} &   \textbf{100 }\\ \bottomrule
\end{tabular}

\vspace{2pt}
\fontsize{9.5pt}{10.25pt}\selectfont
\textsuperscript{*} Results taken from~\cite{MOD++}.
\end{table}
\endgroup


For complex tasks such as ego-motion compensation, deep network architectures are usually required. Table \ref{tab:params} provides a comparison in terms of complexity, by showing the number of parameters of the previously mentioned methods in contrast with the non-learnable parameters of the OMS algorithm. OMS accounts for 80 non-learnable parameters from the weights of the surround ($8 \times 8$) and center ($4 \times 4$) kernels, with radius 4 and 2, respectively. These results illustrate that our method significantly reduces complexity and memory requirements compared to prior approaches while achieving comparable performance. 
\begingroup
\renewcommand{\arraystretch}{1.1}
\begin{table}[h]
\centering
\caption{Comparison of the number of parameters between state-of-the-art methods and our solution.}
\label{tab:params}
\fontsize{10pt}{10pt}\selectfont
\begin{tabular}{lc}
\toprule
Method     & No. Params                 \\
\midrule
EvDodgeNet \cite{EED} & 30k - 3M                   \\
EV-IMO \cite{mitrokhin2020evimo}   & 40k                        \\
SpikeMS $^{*}$  \cite{SpikeMSs}  & 46k \\
GConv \cite{GConv} & 117k                       \\
PointNet++ \cite{pointnet++} & 1.4M                       \\
MSRNN \cite{multi-rnn} & 4.1M                       \\
\textbf{Ours (OMS)} & \textbf{80}                    \\
\bottomrule
\end{tabular}
\end{table}
\endgroup

\subsection{Qualitative assessment}
\noindent
This method exhibits superior performance in scenarios with comparable object and background events in the DVS stream. The opposite cases are those that diminish the mean IoU. This constraint is illustrated in Figure \ref{fig:sequences_evimo}. We denote the ratio between background spikes and moving object spikes as the background-to-foreground ratio ($br$). For sequences with proportional spikes for background movement and moving objects ($br\simeq1$), the OMS output reveals dense regions for object motion, distinguishing them clearly from the background, see wall and table segments in Figure \ref{fig:sequences_evimo}. When the background spike density significantly exceeds the moving objects' density, the OMS algorithm lacks detail. This results in partial object detection or aliasing of object motion with background activity, as many of the mentioned methods do.

\section{Future Works and Discussion}
We present a novel algorithm to extract Object Motion Sensitivity from authentic DVS events. The ablation experiments demonstrates a performance comparable or superior to existing methods in terms of mean IoU and detection rate. Notably, the algorithm significantly reduces complexity, utilizing $\text{10}^\text{3}$ to $\text{10}^\text{6}$ times fewer parameters compared to previous approaches. Furthermore, the bio-inspired approach eliminates the need for training and tuning learnable parameters, making it a domain-agnostic and effective solution for motion segmentation and ego-motion compensation.
This work leverages the insights gained from the neuroscience-inspired software for integration into neuromorphic sensors, such as IRIS \cite{yin2022iris, sinaga2024hardwarealgorithmreengineeringretinalcircuit}. Moreover, incorporating additional retinal functionalities alongside OMS holds immense potential. Embedding multiple features like looming detection, object orientation, and shape extraction could lead to a multi-modal, reliable, and efficient solution for visual odometry tasks in dynamic environments.

\bibliographystyle{splncs04}
\bibliography{egbib}

\begin{thebibliography}{10}
\providecommand{\url}[1]{\texttt{#1}}
\providecommand{\urlprefix}{URL }
\providecommand{\doi}[1]{https://doi.org/#1}

\bibitem{baccus2008retinal}
Baccus, S.A., {\"O}lveczky, B.P., Manu, M., Meister, M.: A retinal circuit that computes object motion. Journal of Neuroscience  \textbf{28}(27),  6807--6817 (2008)

\bibitem{Baden2016RGC}
Baden, T., Berens, P., Franke, K., Rom{\'a}n~Ros{\'o}n, M., Bethge, M., Euler, T.: The functional diversity of retinal ganglion cells in the mouse. Nature  \textbf{529}(7586),  345--350 (Jan 2016). \doi{10.1038/nature16468}, \url{https://doi.org/10.1038/nature16468}

\bibitem{bartolozzi2014flow}
Benosman, R., Clercq, C., Lagorce, X., Ieng, S.H., Bartolozzi, C.: Event-based visual flow. IEEE Transactions on Neural Networks and Learning Systems  \textbf{25}(2),  407--417 (2014). \doi{10.1109/TNNLS.2013.2273537}

\bibitem{motion2007}
Delbruck, T., Lichtsteiner, P.: Fast sensory motor control based on event-based hybrid neuromorphic-procedural system. In: 2007 IEEE International Symposium on Circuits and Systems (ISCAS). pp. 845--848 (2007). \doi{10.1109/ISCAS.2007.378038}

\bibitem{motion2013}
Delbruck, T., Lang, M.: Robotic goalie with 3 ms reaction time at 4

\bibitem{gallego2018unifying}
Gallego, G., et~al.: A unifying contrast maximization framework for event cameras, with applications to motion, depth, and optical flow estimation. In: Proceedings of the IEEE conference on computer vision and pattern recognition. pp. 3867--3876 (2018)

\bibitem{gallego2020event}
Gallego, G., et~al.: Event-based vision: A survey. IEEE transactions on pattern analysis and machine intelligence  \textbf{44}(1),  154--180 (2020)

\bibitem{Hayes:12}
Hayes, T.: Next-generation cell phone cameras. Opt. Photon. News  \textbf{23}(2),  16--21 (Jan 2012). \doi{10.1364/OPN.23.2.000016}, \url{https://www.optica-opn.org/abstract.cfm?URI=opn-23-2-16}

\bibitem{khan2017ego}
Khan, N.H., Adnan, A.: Ego-motion estimation concepts, algorithms and challenges: an overview. Multimedia Tools and Applications  \textbf{76},  16581--16603 (2017)

\bibitem{motion2006}
Litzenberger, M., Posch, C., Bauer, D., Belbachir, A., Schon, P., Kohn, B., Garn, H.: Embedded vision system for real-time object tracking using an asynchronous transient vision sensor. In: 2006 IEEE 12th Digital Signal Processing Workshop \& 4th IEEE Signal Processing Education Workshop. pp. 173--178 (2006). \doi{10.1109/DSPWS.2006.265448}

\bibitem{ma20223d}
Ma, X., Ouyang, W., Simonelli, A., Ricci, E.: 3d object detection from images for autonomous driving: a survey. arXiv preprint arXiv:2202.02980  (2022)

\bibitem{EED}
Mitrokhin, A., Fermüller, C., Parameshwara, C., Aloimonos, Y.: Event-based moving object detection and tracking. In: 2018 IEEE/RSJ International Conference on Intelligent Robots and Systems (IROS). pp.~1--9 (2018). \doi{10.1109/IROS.2018.8593805}

\bibitem{GConv}
Mitrokhin, A., Hua, Z., Fermüller, C., Aloimonos, Y.: Learning visual motion segmentation using event surfaces. In: 2020 IEEE/CVF Conference on Computer Vision and Pattern Recognition (CVPR). pp. 14402--14411 (2020). \doi{10.1109/CVPR42600.2020.01442}

\bibitem{mitrokhin2020evimo}
Mitrokhin, A., Ye, C., Fermuller, C., Aloimonos, Y., Delbruck, T.: Ev-imo: Motion segmentation dataset and learning pipeline for event cameras (2020)

\bibitem{lveczky2003SegregationOO}
{\"O}lveczky, B.P., Baccus, S.A., Meister, M.: Segregation of object and background motion in the retina. Nature  \textbf{423},  401--408 (2003), \url{https://api.semanticscholar.org/CorpusID:4385091}

\bibitem{SpikeMSs}
Parameshwara, C.M., Li, S., Fermüller, C., Sanket, N.J., Evanusa, M.S., Aloimonos, Y.: Spikems: Deep spiking neural network for motion segmentation. In: 2021 IEEE/RSJ International Conference on Intelligent Robots and Systems (IROS). pp. 3414--3420 (2021). \doi{10.1109/IROS51168.2021.9636506}

\bibitem{MOD++}
Parameshwara, C.M., Sanket, N.J., Gupta, A., Ferm{\"{u}}ller, C., Aloimonos, Y.: {MOMS} with events: Multi-object motion segmentation with monocular event cameras. CoRR  \textbf{abs/2006.06158} (2020), \url{https://arxiv.org/abs/2006.06158}

\bibitem{motion2012}
Piątkowska, E., Belbachir, A.N., Schraml, S., Gelautz, M.: Spatiotemporal multiple persons tracking using dynamic vision sensor. In: 2012 IEEE Computer Society Conference on Computer Vision and Pattern Recognition Workshops. pp. 35--40 (2012). \doi{10.1109/CVPRW.2012.6238892}

\bibitem{pointnet++}
Qi, C., Yi, L., Su, H., Guibas, L.J.: Pointnet++: Deep hierarchical feature learning on point sets in a metric space. In: Neural Information Processing Systems (2017)

\bibitem{simulator_MOD}
Sabzevari, R., Scaramuzza, D.: Multi-body motion estimation from monocular vehicle-mounted cameras. IEEE Transactions on Robotics  \textbf{32}(3),  638--651 (2016). \doi{10.1109/TRO.2016.2552548}

\bibitem{MOD}
Sanket, N.J., et~al.: Evdodgenet: Deep dynamic obstacle dodging with event cameras. In: 2020 IEEE International Conference on Robotics and Automation (ICRA). pp. 10651--10657. IEEE (2020)

\bibitem{schwartz2021object}
Schwartz, G.W., Swygart, D.: Object motion sensitivity. In: Retinal Computation, pp. 230--244. Elsevier (2021)

\bibitem{sinaga2024hardwarealgorithmreengineeringretinalcircuit}
Sinaga, J., Clerico, V., Kaiser, M.A.A., Snyder, S., Lohia, A., Schwartz, G., Parsa, M., Jaiswal, A.: Hardware-algorithm re-engineering of retinal circuit for intelligent object motion segmentation (2024)

\bibitem{snyder2021thor}
Snyder, S.E., Husari, G.: Thor: A deep learning approach for face mask detection to prevent the covid-19 pandemic. In: SoutheastCon 2021. pp.~1--8. IEEE (2021)

\bibitem{stroffregen2019}
Stoffregen, T., Gallego, G., Drummond, T., Kleeman, L., Scaramuzza, D.: Event-based motion segmentation by motion compensation. In: 2019 IEEE/CVF International Conference on Computer Vision (ICCV). pp. 7243--7252 (2019). \doi{10.1109/ICCV.2019.00734}

\bibitem{stroffregen2018flow}
Stoffregen, T., Kleeman, L.: Simultaneous optical flow and segmentation {(SOFAS)} using dynamic vision sensor. CoRR  \textbf{abs/1805.12326} (2018)

\bibitem{vasco2017icub}
Vasco, V., Glover, A., Mueggler, E., Scaramuzza, D., Natale, L., Bartolozzi, C.: Independent motion detection with event-driven cameras. In: 2017 18th International Conference on Advanced Robotics (ICAR). pp. 530--536 (2017). \doi{10.1109/ICAR.2017.8023661}

\bibitem{saliency}
Wang, X., Ning, C., Xu, L.: Spatiotemporal saliency model for small moving object detection in infrared videos. Infrared Physics \& Technology  \textbf{69},  111--117 (2015). \doi{https://doi.org/10.1016/j.infrared.2015.01.018}

\bibitem{receptive2018Fields}
Wienbar, S., Schwartz, G.: The dynamic receptive fields of retinal ganglion cells. Progress in Retinal and Eye Research  \textbf{67} (06 2018). \doi{10.1016/j.preteyeres.2018.06.003}

\bibitem{yin2022iris}
Yin, Z., Kaiser, M.A.A., Camara, L.O., Camarena, M., Parsa, M., Jacob, A., Schwartz, G., Jaiswal, A.: Iris: Integrated retinal functionality in image sensors. Frontiers in Neuroscience  \textbf{17} (2023). \doi{10.3389/fnins.2023.1241691}

\bibitem{multi-rnn}
Zhang, S., Sun, L., Wang, K.: A multi-scale recurrent framework for motion segmentation with event camera. IEEE Access  \textbf{11},  80105--80114 (2023). \doi{10.1109/ACCESS.2023.3299597}

\bibitem{zhou2022computer}
Zhou, L., Zhang, L., Konz, N.: Computer vision techniques in manufacturing. IEEE Transactions on Systems, Man, and Cybernetics: Systems  (2022)

\bibitem{zhou2017unsupervised}
Zhou, T., Brown, M., Snavely, N., Lowe, D.G.: Unsupervised learning of depth and ego-motion from video. In: Proceedings of the IEEE conference on computer vision and pattern recognition. pp. 1851--1858 (2017)

\bibitem{zhu2019unsupervised}
Zhu, A.Z., et~al.: Unsupervised event-based learning of optical flow, depth, and egomotion. In: Proceedings of the IEEE/CVF Conference on Computer Vision and Pattern Recognition. pp. 989--997 (2019)

\end{thebibliography}

\end{document}